\documentclass[letterpaper, 10 pt, conference]{ieeeconf}  % Comment this line out if you need a4paper

\IEEEoverridecommandlockouts                              % This command is only needed if 
\usepackage[left=0.75in,right=0.75in,top=0.82in,bottom=0.75in]{geometry}
\usepackage{afterpage}

\usepackage{gensymb}
\usepackage{subfig}
\usepackage{times}
\usepackage{epsfig}
\usepackage{graphicx}
\usepackage{comment}
\usepackage{xcolor}
\usepackage{hyperref}
\hypersetup{
    colorlinks=true,
    linkcolor=black,
    filecolor=black,      
    urlcolor=black,
}
\usepackage[font=small]{caption}
\usepackage{array}
\usepackage{xcolor}
\usepackage{scrextend}
\usepackage[subtle, title]{savetrees}

\newcommand{\ct}[1]{\fontsize{7pt}{1pt}\selectfont{#1}}

\usepackage{multicol}
\usepackage{amsmath}

\title{\LARGE \bf What's in my Room? Object Recognition on Indoor Panoramic Images
}

\author{Julia Guerrero-Viu$^{*1}$, Clara Fernandez-Labrador$^{*1,2}$, C{\'e}dric Demonceaux$^{2}$, and Jose J. Guerrero$^{1}$%
\date{}% <-this % stops a space
\thanks{$^{*}$ Equal contribution}%
\thanks{$^{1}$J. Guerrero-Viu, C. Fernandez-Labrador and J.J. Guerrero. Instituto de Investigaci{\'o}n en Ingenier{\'i}a de Arag{\'o}n (I3A), Universidad de Zaragoza, Spain
        {\tt\footnotesize }}%
\thanks{$^{2}$C. Fernandez-Labrador and C{\'e}dric Demonceaux. VIBOT ERL CNRS 6000, ImViA, Universit{\'e} Bourgogne Franche-Comt{\'e}, France.
        {\tt\footnotesize }}%
}

\begin{document}
\maketitle
\begin{abstract}
In the last few years, there has been a growing interest in taking advantage of the $\textbf{360}^\circ$ panoramic images potential, while managing the new challenges they imply.
%There has been a growing interest in estimating 3D from $360^\circ$ panoramic images. 
While several tasks have been improved thanks to the contextual information these images offer, object recognition in indoor scenes still remains a challenging problem that has not been deeply investigated. This paper provides an object recognition system that performs object detection and semantic segmentation tasks by using a deep learning model adapted to match the nature of equirectangular images. 
From these results, instance segmentation masks are recovered, refined and transformed into 3D bounding boxes that are placed into the 3D model of the room.
Quantitative and qualitative results support that our method outperforms the state of the art by a large margin and show a complete understanding of the main objects in indoor scenes.

\end{abstract}

\section{Introduction} 
% ¿Qué queremos vender?: 
% - Hemos adaptado blitznet a imagenes panoramicas (input size y proposals) y añadido data augmentation (importante ya que tenemos pocos datos) y equiconvs para mejorar el modelo (la mejora es mayor que con layouts). 
% - Hemos añadido un post-processing para hacer instance segmentation (referencias?). 
% - Hemos analizado como se podría beneficiar esta tarea de conocer el layout y hemos visto que se puede mejorar la segmentacion y tener una idea inicial de los objetos en 3D solo a partir de mascaras.
% - Nos hemos comparado con el estado del arte y les ganamos.

%\todo{hay que contestar estas preguntas principalmente: Qué problema queremos resolver (object detection) y por qué es importante resolverlo, es decir, para qué es útil (aplicaciones robotica). Qué chanllenges hay (viewpoint changes, occlusions, illumination variations, background clutter or sensor noise). Cómo lo resolvemos nosotros, nuestra propuesta.}

The increasing interest in autonomous mobile systems, like drones, robotic vacuum cleaners or assistant robots, makes detection and recognition of objects in indoor environments a very important and demanded task. 

%Since recognizing a visual concept is relatively trivial for a human to perform, it is worth considering the hard challenges inherently involved. Objects in images can be oriented in many different ways with respect to the camera and can vary their size, both in terms of scale in the image and their real world size. They can also be occluded, blended into the environment because of their color or appearance, or affected by different illumination conditions, which change drastically their aspect on the pixel level. Moreover, the concept behind an object's name is sometimes relatively broad, having many different types and appearances that could be described under the same concept, including subjective and non-clear frontiers to other concepts. For example, where do you consider the limits between a sofa and an armchair?.

Since recognizing a visual concept is relatively trivial for a human, it is worth considering the hard challenges inherently involved. Objects in images can be oriented in many different ways, vary their size, be occluded, blended into the environment because of their color or appearance, or affected by different illumination conditions, which changes drastically their aspect on the pixel level. Moreover, the concept behind an object's name is sometimes broad, including non-clear frontiers to other concepts. For example, where do you consider the limits between a sofa and an armchair?

Convolutional Neural Networks (CNNs) have already demonstrated to be the best known models to perform object recognition, as they are capable of dealing with those challenges by automatically learning objects' inherent features and correctly identify their intrinsic concepts.

However, images from conventional cameras have a small field of view, much smaller than human vision, which implies that contextual information cannot be as useful as it should. To overcome this limitation, a real impact came with the arrival of the $360^\circ$ full-view panoramic images, which are recently arising more and more interest in the robotics and computer vision community, as they allow us to visualize, in a single image, the whole scene at the same time. Together with all of their potential we have to deal with challenges produced by their own spherical projection, such as distortion, or the lack of complete, labeled and massive datasets. This requires the development of specific techniques that take advantage of their strengths and allow working with panoramic images in an efficient and effective way.

\begin{figure}
\begin{center}
   \subfloat{\includegraphics[width=1\linewidth]{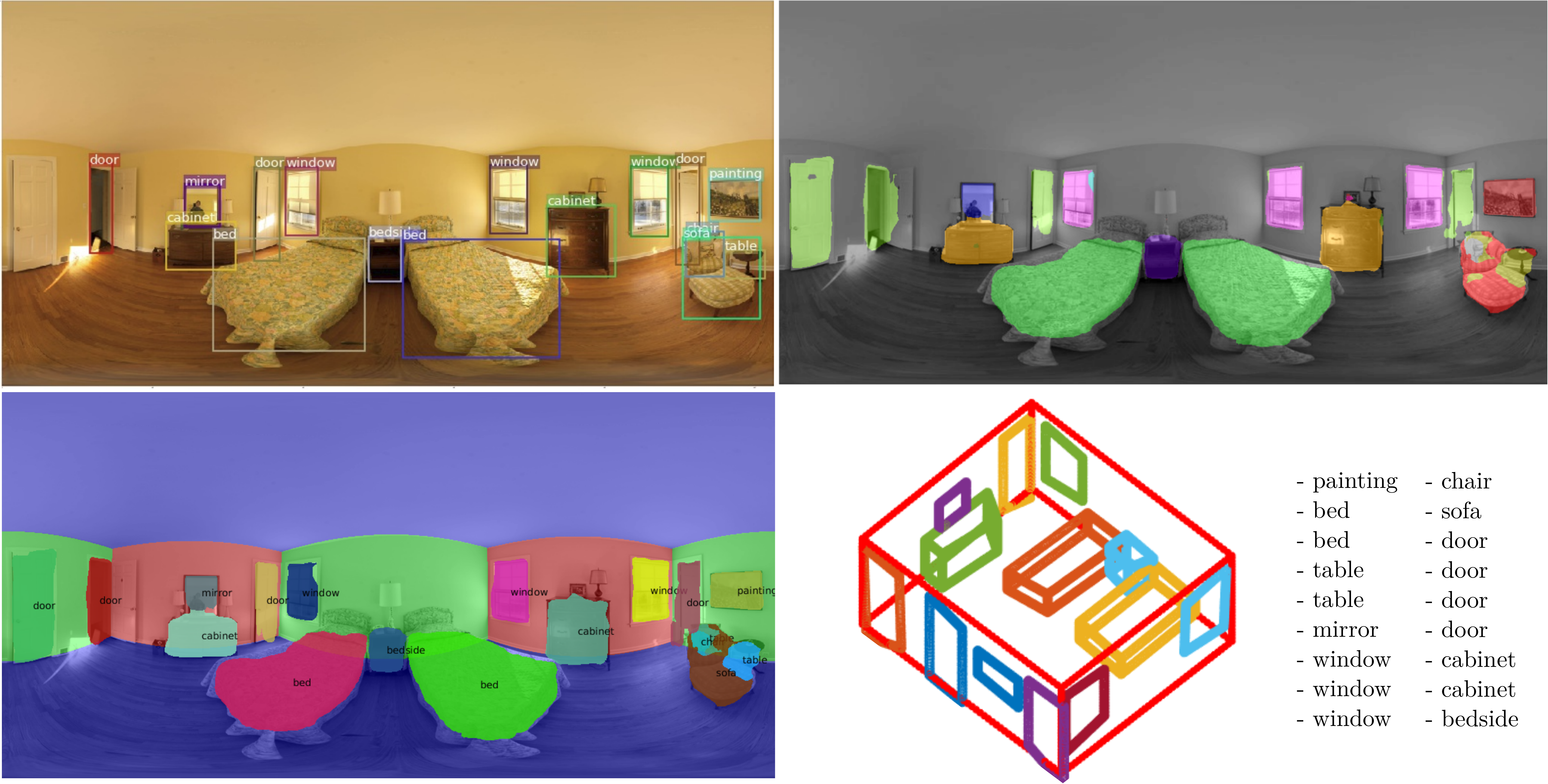}}
\end{center}
   %\caption{Combined information}
   \caption{\textbf{What's in my room?}: from a single spherical RGB image, we obtain localization and pixel-wise classification of the objects present in the scene. We use this information to recover instance segmentation masks of all objects and finally place them inside the 3D room layout.}
   \label{fig:teaser}
\end{figure}

In this paper, we propose an object recognition system that provides a complete understanding of the main objects in an indoor scene from a single $360^\circ$ image in equirectangular projection. Our method extends the BlitzNet model~\cite{BlitzNet} to perform both object detection and semantic segmentation tasks but adapted to match the nature of the equirectangular image input. We train the network to predict 14 different classes of main indoor scenes related objects. Results of the CNN are post-processed to obtain instance segmentation masks, which are successfully refined by taking advantage of the spatial contextual clues that the room layout provides. In this work, we not only show the potential of exploiting the 2D room layout to improve the instance segmentation mask, but also the possibility of leveraging the 3D layout to generate 3D object bounding boxes directly from the improved masks.

%In this paper, we propose an object recognition system that allows to get a complete understanding of the main objects from indoor panoramic images. It is composed by an extension of BlitzNet \cite{BlitzNet} to perform both object detection and semantic segmentation tasks but adapted to match the nature of the equirectangular image input, and a post-processing model that shows the potential of exploiting the 3D layout to improve segmentation and generate 3D object bounding boxes.

%We feed a single panoramic RGB image to the input convolution channels. We train the network to predict 14 different classes of main indoor scenes related objects, both for detection and segmentation. By post-processing the network output, we go forward and create instance segmentation masks, which are successfully refined by taking advantage of the room layout. Finally, \todo{queda añadir lo que haces con el 3D que mejor que lo redactes tu :)}

%The network architecture follows the meta-architecture of BlitzNet adapting their propwith architecture adaptation and tuning to match the nature of the equirectangular image input. (\todo{summary of adaptation JULIA: yo no diria que adaptamos la arquitectura, la adaptacion es basicamente lo de las proposals y q use nuestro dataset, tampoco la explicaria mas alla}). 

\section{Previous Work}
Object detection field has been mainly dominated by two different approaches: one-stage and two-stage detectors. Two-stage detectors, as the first R-CNN\cite{RCNN} architecture followed by its variants Fast R-CNN\cite{FastRCNN}, Faster R-CNN\cite{FasterRCNN} and Mask R-CNN\cite{MaskRCNN} achieve great accuracy but lower speed. They require firstly to refine proposals to obtain the features needed to classify the objects. On the other hand, one-stage detectors, following YOLO\cite{YOLOv2} and SSD\cite{SSD} simultaneous bounding box refinement and classification, significantly reduce computational cost. They achieve real-time performing maintaining high accuracy, which is needed for most applications in autonomous mobile systems. SSD multi-scale pyramid idea proves to help in conducting more accurate detections and manage widely various object sizes, approach followed in most state-of-the-art object detectors. 

While all those models optimize bounding box detection, not so many integrate in their pipeline the pixel-wise recognition needed for many applications. In this way, BlitzNet~\cite{BlitzNet} is a one-stage multi-scale model that adds semantic segmentation and therefore recognizes objects at pixel level. It also proves the advantages of jointly learning two scene understanding tasks: object detection and semantic segmentation, which benefit from each other by sharing almost the complete network architecture.

However, state-of-the-art research mainly focuses on using conventional images. Their limited field of view prevents contextual information from being as crucial as it is in scene understanding for humans. Differently from outdoor object recognition, where thanks to the increasing research on autonomous driving, there are recent works using panoramic images~\cite{traffic}~\cite{more_traffic}, there is no wide research on object recognition from indoor panoramas. A recent work that addresses this problem is~\cite{Deng}, where Deng \textit{et. al} use a R-CNN approach, and also evaluate their own implementation of DPM~\cite{DPM} on panoramas. The most relevant work on indoor panoramic object recognition is PanoContext~\cite{PanoContext}. It includes 2D object detection and semantic segmentation among other 3D scene understanding tasks, proving the potential of having a larger field of view for recognition problems. Their method, nevertheless, is based on geometrical reasoning and traditional computer vision feature extractors and can be still considered as state-of-the-art in indoor object recognition on panoramic images. 
Recent research on this kind of images includes 3D layout recovery~\cite{omedes2013omnidirectional}~\cite{jia2015estimating}~\cite{LayoutNet}~\cite{fernandez2019corners} and scene modeling~\cite{depth_maps}, which provides global context and gives a 3D interpretation of the scene from a single view. In~\cite{lukierski2017room}, they show that  this tasks can also benefit and augment an omnidirectional SLAM. Combining object recognition and 3D layout recovery motivates our proposal to obtain the 3D recognition and location of main objects in our room.

\begin{comment}
\subsection{Object Recognition}
Talk a little about object recognition with conventional images, maybe two-stages versus one-stage.
BlitzNet~\cite{BlitzNet} (talk here about their model Fully Convolutional, ResSkip modules, joint learning and their experiments proving it and multi-scale recognition).

\subsection{Panoramic Images}
PanoContext~\cite{PanoContext} talk about their demonstrations of the field of view and the usefullness of context in object detection. Their results in object detection and layout recovery. Be clear that there are no many works on object recognition on panoramic indoor scenes. Maybe mentioned some outdoors works and then other tasks such as layout recovery, CFL~\cite{fernandez2019corners} and EquiConvs, and maybe depth maps.
\end{comment}

\section{Dataset extension}
% Vender un poco mas la falta de datasets de panoramicas completos y que por eso hemos extendido el SUN360 nosotros

%For this work, SUN360 (Scene UNderstanding $360\degree$ panorama) database~\cite{SUN360} is used for training, validation and evaluation of the system. Bedroom and living room sets, formed by 418 and 248 images respectively and 14 different object classes are considered. Dataset is divided into 85\% for train and validation and 15\% for test. 2D bounding boxes labels are taken from PanoContext~\cite{PanoContext} work, and segmentation masks are automatically own-generated as detailed below. The complete \textbf{dataset} used in this work has already been released for public access and can be found here: \href{https://cfernandezlab.github.io/room_OR/}{project webpage}.
Panoramic images datasets with object recognition labels are not as standard or complete as conventional images ones~\cite{SUN360}~\cite{SUNCG}~\cite{Standford2D}. Therefore, in this work we decide to extend the SUN360 database~\cite{SUN360} with segmentation labels. For every panorama, we generate individual masks encoding each object's spatial layout. Additionally, we combine all the masks obtaining a semantic segmentation panoramic image with per-pixel classification.  %without differentiating instances. %Therefore, in this work we decide to extend SUN360 database~\cite{SUN360} with 2D segmentation maps, creating both binary masks for each of the objects and semantic segmentation maps of the full image. 

Bedroom and living room sets, formed by 418 and 248 images respectively, are used and 14 different object classes are considered. The dataset is divided into 85\% for train and validation and 15\% for test.

\begin{figure}
\begin{center}
   \subfloat{\includegraphics[width=0.95\linewidth]{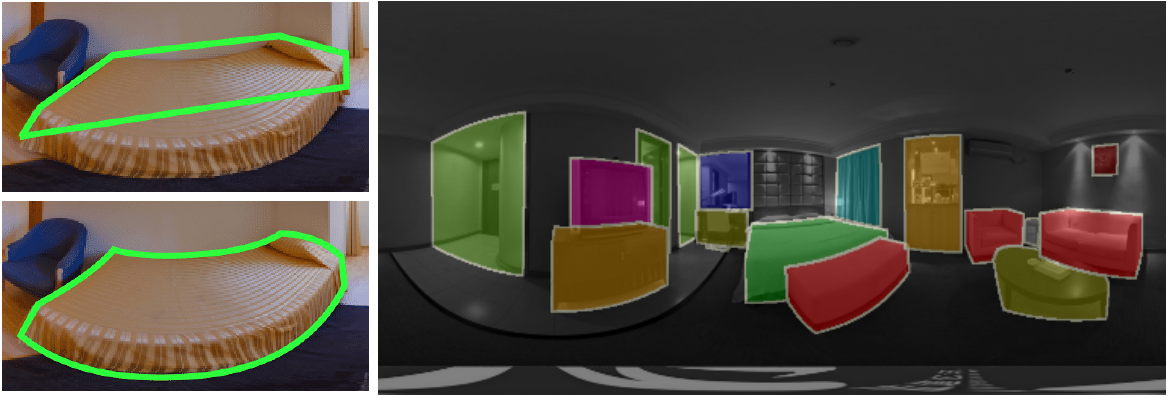}}
\end{center}
   \caption{Result of our method to \textbf{create semantic segmentation masks}, assuming hypothesis of occlusion. Notice on the left the differences between creating straight contours on image domain (top) vs. spherical domain (bottom).}
   \label{fig:dataset_masks}
\end{figure}

We generate segmentation masks based on 2D bounding points of the objects, taken from PanoContext~\cite{PanoContext} work. We project them on the spherical domain to follow distortion patterns in contours and to correctly manage objects that appear cropped on the horizontal image limits. To combine the binary masks and create the semantic segmentation panorama, with the lack of depth or other 3D information, an hypothesis of occlusion among objects is needed. We consider the assumption that objects are not in general completely occluded, and therefore for each pair of objects in conflict their area of overlap and size are computed. If area of overlap is bigger than a threshold, the smallest object is considered closer and completely visible and otherwise the biggest one is selected. With its evident limitations, this hypothesis experimentally proves to work well in most of the cases, allowing to correctly segment most of the visible and cuboid-shaped objects in images as shown in Figure~\ref{fig:dataset_masks}.
\vspace{1.0cm}

The complete \textbf{dataset} used in this work is released for public access and can be found in the project webpage\footnote{\label{web}Available at \href{https://webdiis.unizar.es/~jguerrer/room_OR/}{\color{blue}https://webdiis.unizar.es/$\sim$jguerrer/room\_OR/}
    {\tt\footnotesize }}. 

\section{Model}
In this section we present our object recognition model, called \textit{Panoramic BlitzNet}, that is based on the original CNN BlitzNet~\cite{BlitzNet} but adapted to work specifically with complete equirectangular images. It addresses both object detection and semantic segmentation tasks, following BlitzNet architecture: a Fully Convolutional model that follows the encoder-decoder approach with skip connections. It performs multi-scale recognition and takes advantage of joint learning. Main changes to their base implementation include the use of the complete rectangular panorama, modifying the input aspect ratio. We also change the anchor boxes proposals, as the new input shape needs to be considered because they are centered on pixels grid. Our bounding boxes proposals are done by firstly converting image to a regular grid, covering the whole rectangular-shaped image. Grid has different dimensions in each layer, from 128x256 to 1x2, because of the iteratively lower scale of the feature maps. In each grid cell 5 different proposals are created with 5 different aspect ratios: 1, 2, 1/2, 3 and 1/3, allowing the network to manage different object shapes. 

Special mention deserves data augmentation as an important technique to avoid overfitting, particularly on non-massive datasets like in our case. Here, we modify the original data augmentation by removing random crops on images (contextual information is important) and adding horizontal rotation from $0\degree$ to $360\degree$ to cover all different positions on the sphere.

\subsection{How can we deal with $360^\circ$ images distortion?}
We exploit the potential of omnidirectional images covering $360^\circ$ horizontal and $180^\circ$ vertical field of view represented in equirectangular projection. While these images allow us to analyse the whole scene at once taking advantage of all the context, they present great distortions due to their projection of the sphere. 
Here, we replace all standard convolutions of our \textit{Panoramic BlitzNet} by equirectangular convolutions (EquiConvs \cite{fernandez2019corners}), to study their impact on the task of recognizing objects. With this kind of convolutions, the kernel \textbf{adapts its shape and size accordingly to the distortions} produced by the equirectangular projection.
As mentioned in \cite{fernandez2019corners}, the distortion presented is location dependent, specifically, it depends on the polar angle. They demonstrate how EquiConvs can be really convenient to generalize to different camera positions since the layout shape can suffer from many variations. For the specific task of object recognition, the use of EquiConvs is definitely convenient even if the camera is always at the same place, since objects can be at many different locations inside the scene --\textit{e.g.} objects closer to the camera will have greater distortions than objects around the horizon line. EquiConvs here play an important role since they can learn ignoring this distortion patterns and thus, being more able to learn real objects appearance.
Additionally, one important challenge to accomplish our goal is represented by the need of extensive annotations for training object recognition. To this end, EquiConvs make the \textbf{pre-training much more effective} since this type of convolutions implicitly handle equirectangular distortion, being able to use previous weights from conventional images as if they were learnt on the same kind of images. We can therefore exploit the wealth of publicly available perspective datasets for training -- SUN RGB-D~\cite{SUNRGBD} dataset in this case, which reduces the cost of annotations and allows training under a larger variety of scenarios.
%Additionally, without panoramic massive datasets, our pre-training is highly necessary and EquiConvs make it much more effective: weights on conventional images are learnt without distortion and EquiConvs implicitly handle equirectangular distortion, being able to use previous weights as if they were learnt on the same kind of images. 
Moreover, standard convolutions do not understand that the image wraps around the sphere, loosing the continuity of the scene, while EquiConvs, working directly on the spherical space, avoid padding and exploit this idea. This makes them also very suitable for the objects that appear cut between the left and right side of the image.

%Apart from the modified model to be fed with panoramic images, we have studied the impact of equirectangular convolutions (EquiConvs~\cite{CFL}) on the task of object recognition. We integrate this kind of convolutions on the system, making our \textbf{pre-training much more effective}: Weights on conventional images are learnt without distortion and EquiConvs implicitly handle equirectangular distortion, being able to use previous weights as if they were learnt on the same kind of images. Additionally, taking into account that our dataset does not include camera pose changes, there is a risk that standard convolutions overfit its distortion patterns, which does not happen when using EquiConvs that are invariant to those changes, as already proved in~\cite{CFL}. This capacity of \textbf{avoiding overfitting} makes them very promising for generalization on very different datasets.

\subsection{From semantic to instance segmentation}
Semantic segmentation masks allow us to pixel-wise classify scenes in object categories. One step further goes instance segmentation, which classifies each pixel not only to its category but also differentiating its concrete object instance, an essential stage to correctly locate them into the 3D reconstruction of the room. Without using any instance segmentation prior to be learnt, we add a simple post-processing to obtain \textbf{pixel-wise instance classification} of the scene, based on the outputs of the network. Considering each bounding box detection as a Gaussian distribution, it is assumed that 99\% of the object is contained on it, so standard deviation in each dimension is taken as $\sigma_w = \frac{width}{6}, \sigma_h = \frac{height}{6}$, giving $3\sigma$ at each side of the mean, which is defined as the center of the bounding box. Then, each pixel from the semantic segmentation mask is assigned to the instance distribution with the minimum Mahalanobis distance. When none of the distances to distributions exceeds the chi-squared test threshold, the pixel is left as the classification in the initial semantic segmentation, a proposal that shown the best experimental results.

This assumption gives, consequently, more importance to bigger objects, which are usually better segmented by our model. In addition to providing an approach to create instance segmentation masks, we analyze the impact it can have in improving our initial segmentation, as shown in experiments. 

\subsection{Can we convert instance segmentation masks into 3D bounding boxes?} 
If there is a task that has experimented a disruptive innovation with the emergence of $360^\circ$ images, it has been the room layout estimation problem \cite{jia2015estimating,LayoutNet,fernandez2018layouts,fernandez2019corners,sun2019horizonnet}. In these works, from a single panorama, they recover the main structure of the room --\textit{i.e.} disposition of the walls, ceiling and floor, not only in the image domain, but also a complete 3D reconstruction model up to scale.

The intuition is that objects location and pose inside a 3D indoor space are not randomly distributed. Following the law of physics, objects will be fairly constrained to lie on at least one supporting plane, in stable configurations and, in several cases, aligned with the room walls. 
This means that the room layout provides strong spatial contextual clues as to where and how objects can be found. 
%If there is a task that has experimented a disruptive innovation with the emergence of $360^\circ$ images, it has been the room layout estimation problem \cite{LayoutNet,fernandez2018layouts,fernandez2019corners,sun2019horizonnet}. In recent years we have seen a growing interest in solving this problem taking advantage of all the context offered by this type of images, obtaining an outstanding performance and understanding compared with traditional methods using conventional images \cite{RoomNet, dasgupta2016delay, zhao2017physics}. 

Thus, in order to provide a greater understanding of the scene, here we analyze the potential of using the room layout as a
\textit{prior} combined with the object recognition task:
%and the far-reaching effects of taking into account both the 2D and the 3D layout of the room.

\noindent1) We found that we can easily leverage the room layout in the image domain to \textbf{improve the instance segmentation masks}. Based on the contextual information given by the layout, there are a series of logical assumptions that we can immediately make --\textit{e.g.} it is very unlikely to find doors not resting on the floor or paintings hanging in between two walls. During this process we also detect holes in masks and fill them.

\noindent2) The aforementioned methods provide 3D layout models of the rooms. This allows us to \textbf{place the identified objects inside the 3D model of the room} as long as they lie on the walls, or rest on the floor / ceiling aligned with the walls. In this way, only by detecting the masks of the objects and with a good layout prior, we can obtain a very precise 2D representation of the objects and an initial estimate of the 3D understanding of the scene.

To obtain the room layout, here we choose the work of Fernandez \textit{et al.} \cite{fernandez2019corners}. Additionally, we will need to assume Manhattan World \cite{coughlan1999manhattan} whereby there exist three dominant orthogonal directions defining the scene. Here, we compute the vanishing points of the scene following the approach of \cite{fernandez2018layouts}. They propose a RANSAC-based algorithm that
works directly on omnidirectional images running up to 5 times faster than other approaches. 

\begin{figure}
\begin{center}
   \subfloat{\includegraphics[width=0.95\linewidth]{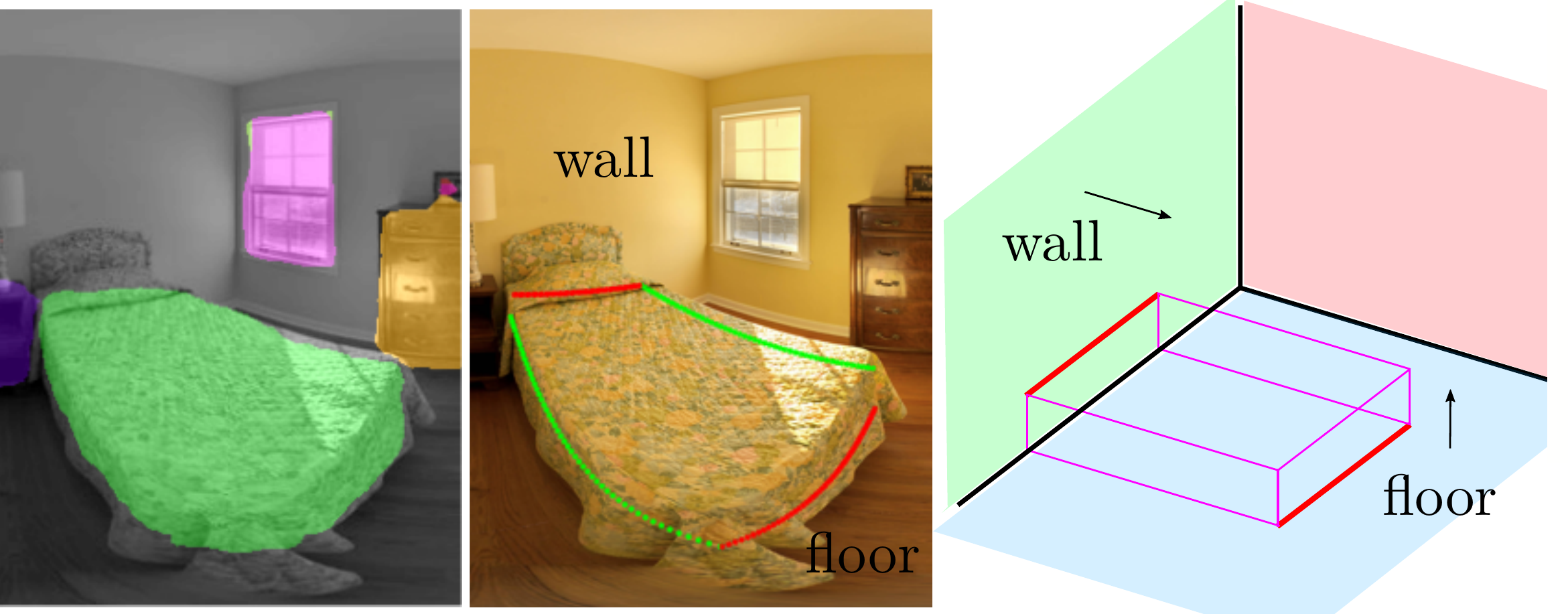}}
\end{center}
   \caption{\textbf{From mask to 3D}: We find the lines that best fit the object mask by a RANSAC algorithm and orient them accordingly to the main directions of the scene. The object dimensions are obtained trusting in the line resting on the wall and the line resting on the floor (red lines in the figure).}
   \label{fig:3dobjects}
\end{figure}

\begin{figure*}
\begin{center}
   \subfloat{\includegraphics[width=1\linewidth]{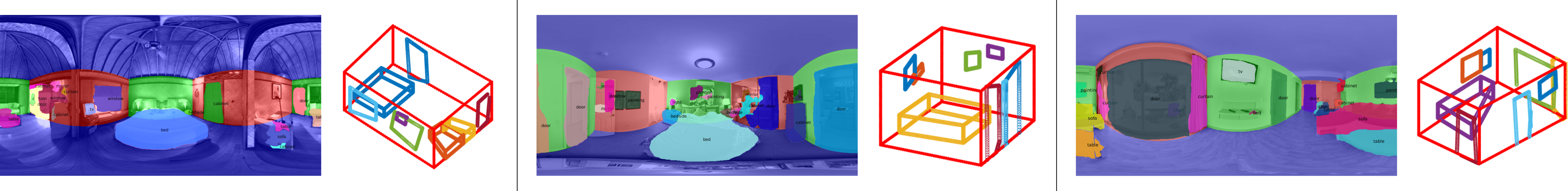}}
\end{center}
   \caption{Examples of 3D models obtained from instance objects masks and room layout knowledge \cite{fernandez2019corners}.}
   \label{fig:examples3D}
\end{figure*}

For the 3D object recognition task, in PanoContext \cite{PanoContext} they generate many cuboid hypotheses combining two approaches. First, they perform rectangle detection in six axis-aligned views projected from the original panorama. Then, they sample rays from the vanishing points to fit image segmentation boundaries obtained by selective search. Finally, they choose the best cuboid whose projection has the largest intersection over union score with the segment. Here instead, we directly approximate every object mask with four lines by a RANSAC approach and classify these lines accordingly to the vanishing points, vp, --\textit{i.e.} we obtain the orientation of the objects inside the scene. We consider that a line belongs to one concrete direction, $k={x,y,z}$, when its normal direction in the 3D space $n_{i}$  fulfills the condition of perpendicularity with the direction under an angular threshold  of $\pm 0.5^{\circ}$ experimentally obtained, $|arccos(n_{i}\cdot vp_{k})-\frac{\pi}{2}|\leq \theta_{th}$. Then, we determine with which planes of the room each object interacts by obtaining the intersections between the object mask and the 2D room layout segmentation map. The room layout segmentation map is directly generated from the predicted layout corners \cite{fernandez2019corners}, encoding each plane accordingly to the main directions of the scene. If we predict that the object lies on a wall --\textit{i.e.} doors, windows, mirrors, pictures, etc., we simply project the object mask to its position in the wall in the 3D room layout. For cuboid objects like beds, sofas or bedside tables, we can obtain the object dimensions --\textit{i.e.} length, width and height, using the lines interacting with the room planes. Figure~\ref{fig:3dobjects} shows how we deal with these objects. With these ideas, we can place most of the objects inside the 3D scene and obtain a good understanding of the scene from the 2D segmentation masks. Some qualitative results are shown in Figure~\ref{fig:examples3D} as well as in the teaser image, Figure~\ref{fig:teaser}.

\vspace{-3mm}

% Posibles problemas: si el layout no esta bien estimado, podemos empeorar nuestra segmentación en lugar de mejorarla. Como trabajo futuro molaría ver como pueden mejorarse entre ellas, demanera bidireccional y no solo unidireccional. Las lineas que se ajustan a la mascara no siempre ayudan al modelo 3D

% ejemplo de bbox 3D en la imagen?
\vspace{0.5cm}
\section{Experiments}
We evaluate our model by different experiments conducted on SUN360~\cite{SUN360} extended dataset, which are presented in this section. Experimental setup is explained in order to make our work reproducible, together with the detailed evaluation metrics.

\subsubsection*{\textbf{Experimental setup}}

The whole model is coded in Python 3.5 using the framework Tensorflow v1.13.1. All experiments were conducted on a single Nvidia GeForce GTX 1080 GPU.

As in \cite{BlitzNet}, we use ResNet-50 as feature extractor, Adam stochastic algorithm~\cite{Adam} for optimization and learning rate set to $10^{-4}$ and decreased twice during training. Experiments were conducted by changing that learning rate without noticeable influence. We use stride 4 in the last layer of the up-scaling stream and varying mini-batch sizes, which are stated in each experiment. All models are trained until convergence, measured with a random validation subset.

%Based on our dataset characteristics, we decide to use pre-trained models instead of a random initialization that would conduct to a clear overfit to our data, as it is studied in the first experiment. Because of the lack of massive panoramic datasets, conventional images are used for pre-training: Firstly, we use the publicly available weights of ResNet50 backbone on ImageNet~\cite{ILSVRC} dataset. With that initialization, we then train the whole network on SUN RGB-D~\cite{SUNRGBD} dataset, already processed to have the same common classes as SUN360. This way we have an initialization for the complete model to be able to fine-tune it with the panoramic images, which is possible thanks to a Fully Convolutional network where weights can be shared ignoring the input dimensions.
Based on our dataset characteristics, we decide to pre-train our network instead of initializing it randomly, that would conduct to a clear overfit to our data, as studied in the first experiment. Because of the lack of massive panoramic datasets, conventional images are used for pre-training: Firstly, we use the publicly available weights of ResNet50 backbone on ImageNet~\cite{ILSVRC} dataset. With that initialization, we then train the whole network on SUN RGB-D~\cite{SUNRGBD}, pre-processed to have the same common classes. This way we have an initialization for the complete model to be able to fine-tune with the panoramic images, possible thanks to a Fully Convolutional network where weights can be shared with variable input dimensions.

%This metric evaluates interpolated precision at all different recall levels, in its simplest definition, but can also be widely found in literature as weighted by the recall area that they represent. In this paper we mostly use simple $mAP$ and when using the second version it is referred as $mAP^{w}$. 
\subsubsection*{\textbf{Evaluation metrics}}
To evaluate detection performance we use typical mean average precision ($mAP$), considering that a predicted bounding box is correct if its intersection over union with the ground truth is higher than 0.3, as exact localization is better predicted in the segmentation branch. The average precision evaluates interpolated precision at all different recall levels, in its simplest definition, but can also be widely found in literature as weighted by the recall area that they represent. In this paper we mostly use simple $AP$ and when using the second version it is referred as $AP^{w}$.
Finally, when calculating the mean among all different classes we use a weighted mean as defined in equation~\ref{eq:map}, being $M$ the number of classes, $AP_i$ the average precision per class, $d_i$ the number of detections of class $i$ and $n$ the total number of detections. It is considered as a more representative metric because results calculated from objects with a minimum number of samples in the test set should be less significant when analyzing a global performance.

Segmentation performance is measured with mean intersection over union ($mIoU$), as stated in equation~\ref{eq:miou}, being $A_i$ the area formed by all pixels of class $i$ in the ground truth
segmentation map, and $\hat{A_i}$ in the predicted segmentation map.
%(except when comparing with other state-of-the-art methods that use a normal average, for fair comparison) 
%\begin{equation}
%    mAP = \frac{1}{M}\sum_{i=1}^{M}\frac{d_i}{n}AP_i
%\end{equation}

\noindent\begin{minipage}[h]{.45\linewidth}
\begin{equation}
  mAP = \sum_{i=1}^{M}\frac{d_i}{n}AP_i
 \label{eq:map}
\end{equation}
\end{minipage}%
\begin{minipage}[h]{.5\linewidth}
\begin{equation}
  mIoU = \frac{1}{M}\sum_{i=1}^{M}\frac{A_i \cap \hat{A_i}}{A_i \cup \hat{A_i}}
  \label{eq:miou}
\end{equation}
\end{minipage}

\subsection{Initialization}
\label{sec:initialization}
First experiment was conducted before the development of our model, to verify the importance of pre-training. It uses original BlitzNet300 architecture, without modifying the base implementation to adapt for panoramas. Batch size is set to 16 and it is trained until convergence. When training, three different initializations are executed to compare: random initialization (trained all from scratch), ImageNet initialization of the feature extractor (ResNet) and pre-trained SUN RGB-D initialization of the complete model. Results are compared in Table~\ref{table:init}, which shows that pre-training with SUN RGB-D followed by fine-tuning the complete network with panoramic dataset, gives the best performance results and generalizes better. ImageNet initialization converges faster, but it demonstrates higher overfitting than SUN RGB-D pre-training. This experiment shows that given a relatively small panoramic dataset, the use of a massive one of conventional images for pre-training allows the network to learn higher level characteristics of the objects, avoiding overfitting and being one of the keys for the success of the system.

%Results are still generally low, due to the fact that we use original \textit{BlitzNet}.

\begin{table}
\centering
\begin{tabular}{l|rr|rr}
\multicolumn{1}{c|}{\begin{tabular}[c]{@{}c@{}}\end{tabular}} & \multicolumn{1}{c}{\begin{tabular}[c]{@{}c@{}}TRAIN\\ $mAP^w$\end{tabular}} & \multicolumn{1}{c|}{\begin{tabular}[c]{@{}c@{}}TEST\\ $mAP^w$\end{tabular}} & \multicolumn{1}{c}{\begin{tabular}[c]{@{}c@{}}TRAIN\\ $meanIoU$\end{tabular}} & \multicolumn{1}{c}{\begin{tabular}[c]{@{}c@{}}TEST\\ $meanIoU$\end{tabular}} \\ \hline
From scratch & 0.911 & 0.468 & 0.872 & 0.419  \\ \hline
ImageNet (ResNet) & 0.896  & 0.479 & 0.788 & 0.432 \\ \hline
SUN RGB-D & 0.728  & \textbf{0.516} & 0.742 & \textbf{0.461}                 
\end{tabular}
\caption{\small\textbf{Effect of initialization:} Results tested on original BlitzNet with different weights initializations. Notice how initializing with SUN RGB-D weights gives clearly better test results. Evaluation on train set is shown to observe the overfitting effect, specially clear in trained from scratch model.} 
\label{table:init}
\end{table}

\begin{table}
\centering
\begin{tabular}{l|c|c|c}
\multicolumn{1}{c|}{} &
 \multicolumn{1}{c|}{$mAP^w$} & \multicolumn{1}{c|}{$mAP$} &
 \multicolumn{1}{c}{$meanIoU$} \\ \hline
BlitzNet & 0.516 & 0.688  & 0.461 \\ \hline
 \textit{Panoramic BlitzNet}  & \textbf{0.632} & \textbf{0.768} & \textbf{0.530}
\end{tabular}
\caption{\textbf{Effect of adapting the CNN for panoramas:} Comparison between results on panoramic images with BlitzNet vs. our proposed \textit{Panoramic BlitzNet}.}
\label{table:panorama_results}
\end{table}

% ----------- Tabla para que salga bien colocada 
\newcolumntype{x}[1]{>{\centering}p{#1pt}}
\newcolumntype{y}{>{\centering}p{16pt}}
\newcommand{\hl}[1]{\textbf{\underline{#1}}}
\renewcommand{\arraystretch}{1.4}
\setlength{\tabcolsep}{4pt}

\begin{table*}[t]
\begin{center}
\small
\vspace{-1em}
\label{tab:vgg_all}
%\sffamily
\resizebox{\linewidth}{!}{
\begin{tabular}{x{95}|x{20} |yyyyyyyyyyyyyyc}
  \ct{input} & mIoU & \ct{backgrd.} & \ct{bed} & \ct{painting} & \ct{table} & \ct{mirror} & \ct{window} & \ct{curtain} & \ct{chair} & \ct{light} & \ct{sofa} & \ct{door} & \ct{cabinet} & \ct{bedside} & \ct{tv} & \ct{shelf}  \\
  \hline\hline
  Initial seg. map  & 53.0 & 90.7 & 61.7 & \textbf{32.1} & \textbf{75.2} & \textbf{42.3} & 55.8 & 54.0 & 55.1 & 31.4 & \textbf{34.6} & \textbf{63.6} & 48.5 & \textbf{40.7} & 52.2 & 57.4  \\
   Post-processed seg. map  & \textbf{53.1} & 90.7 & \textbf{63.3} & 30.1 & 75.0 & 41.5 & \textbf{56.7} & \textbf{55.3} & \textbf{55.5} & \textbf{34.0} & 31.8 & 62.9 & \textbf{48.8} & 40.2 & \textbf{53.5} & \textbf{57.5} \\
\end{tabular}
}
\caption{Semantic segmentation results before and after applying the \textbf{instance segmentation post-processing}. Initial semantic segmentation is taken from our CNN output.}
\label{table:instance}
\end{center}
\end{table*}
\vspace{-2mm}
% ------------ FIN tabla instance

\subsection{Square vs. Panoramic}
This experiment compares the performance of our \textit{Panoramic BlitzNet} network with the original model designed for conventional images. In this case, batch size is reduced to 4, for memory limitations in our GPU. 
As seen in Table~\ref{table:panorama_results}, our adapted model demonstrates clearly better results, improving performance by a wide margin and supporting our assumption of the important benefits of a concrete model designed for panoramas. 
%Qualitative results also prove this assumptions, as analyzed in Figure~\ref{fig:experiments_seg}.

Apart from avoiding distortions and crops to make it fit to a square shape, it also shows the benefits of using a wider field of view, which allows to consider the whole context of the room. This idea is a strong support for the potential of panoramic images, not only in object recognition but in many other visual tasks, at least in those related to indoor scene understanding problem. 

\subsection{StandardConvs vs. EquiConvs}
\label{sec:stand_equiconvs}
\begin{figure}
\begin{center}
%\hspace{-1cm}
   %\subfloat{\includegraphics[width=1\linewidth]{images/paper_results_2.pdf}}
   \subfloat{\includegraphics[scale=0.85]{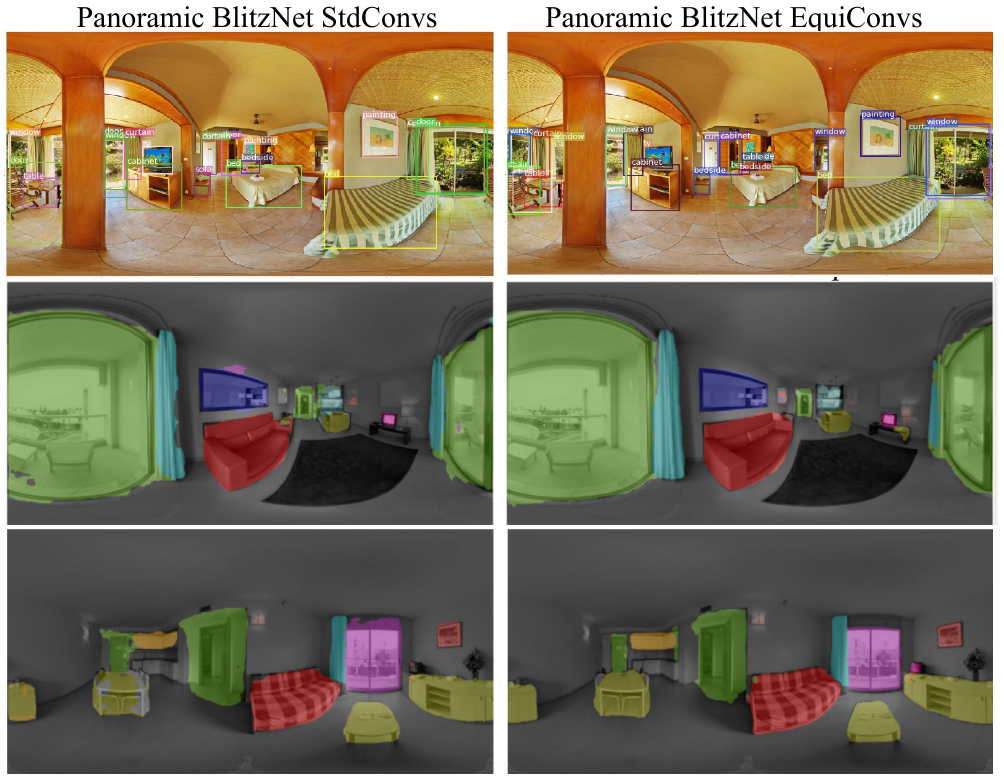}}
\end{center}
   \caption{\textbf{Qualitative evaluation of object detection and semantic segmentation:} Examples of results obtained with our \textit{Panoramic BlitzNet} using both standard convolutions and EquiConvs \cite{fernandez2019corners}.}
    \label{fig:experiments_seg}
\end{figure}

Evaluation of the influence that equirectangular convolutions have on object recognition task is a key point in this work. Comparison of both models can be seen in Tables~\ref{table:state_art_comparison_det} and~\ref{table:state_art_comparison_seg} and Figure~\ref{fig:experiments_seg}. There we show that our model with EquiConvs, adapting the kernel to manage the equirectangular distortion, perform better in both object detection and semantic segmentation tasks. %Differences among categories, however, confirm that this improvement is not very conclusive. Although equirectangular convolutions do not prove to be as beneficial as they intuitively seemed, they still have advantages over standard convolutions. 
Additionally, equirectangular convolutions have other advantages over standard convolutions. They make our pre-training on conventional images more meaningful, as managing distortions by the kernel allows to share the weights as if they were trained on the same kind of images. Therefore, we strongly believe that this type of convolutions help in avoiding overfitting to training data, which due to the particularities of the SUN360 dataset (camera pose does not vary and scenes are relatively similar) does not drastically damage test results, but will probably be crucial when working on different datasets. Finally, EquiConvs also prove to make detections with higher confidence as, when raising the confidence threshold to $0.95$, their recall is maintained over $40\%$ compared to $28\%$ achieved with standard convolutions. Further qualitative results can be found in our project webpage \footref{web}.

% ----------------------------------------------------------
\subsection{Instance segmentation post-processing}
\begin{figure}
   \begin{minipage}[t]{0.24\textwidth}
     \centering
     \includegraphics[width=0.99\linewidth]{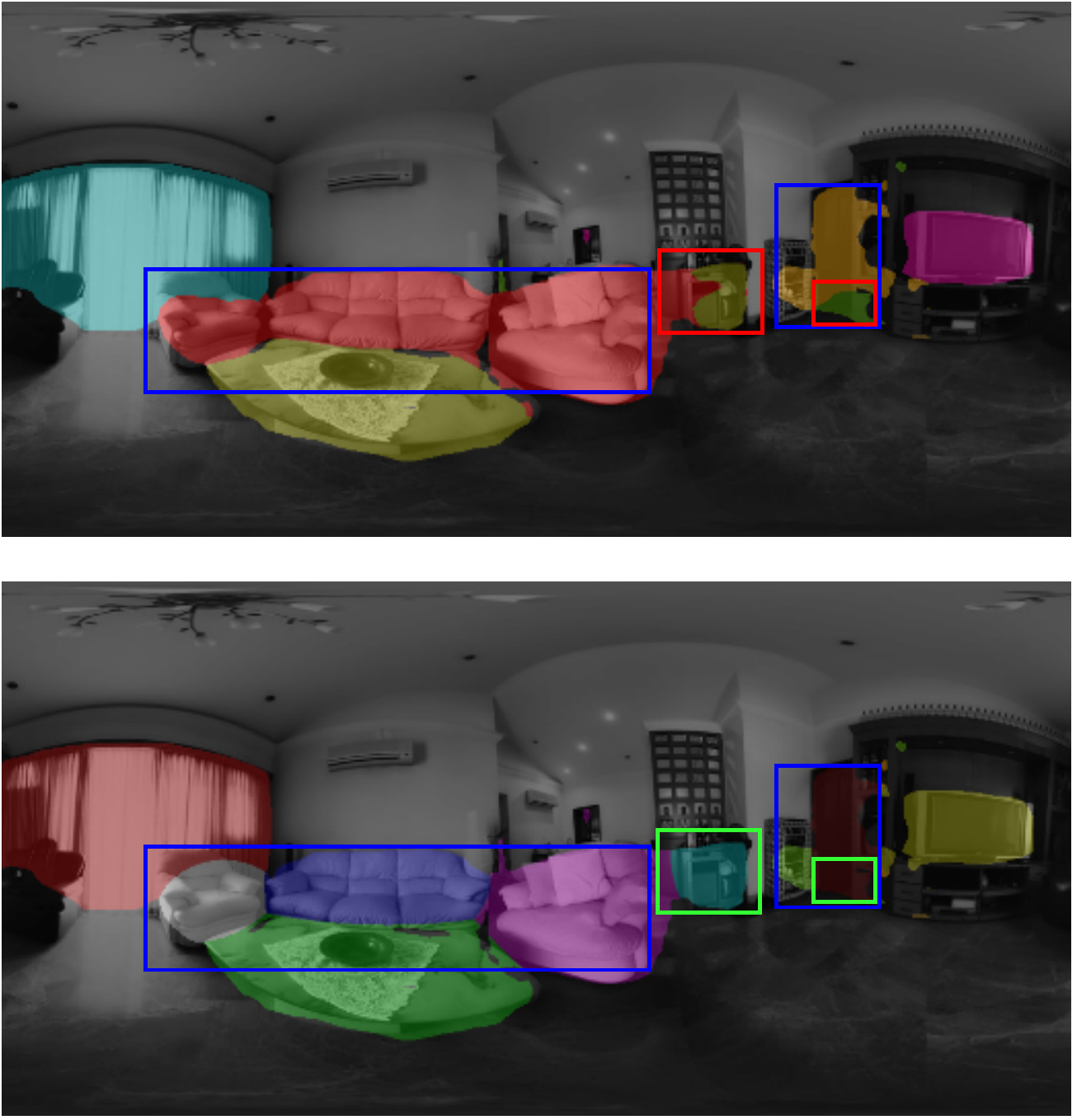}
   \end{minipage}\hfill
   \begin{minipage}[t]{0.24\textwidth}
     \centering
     \includegraphics[width=0.99\linewidth]{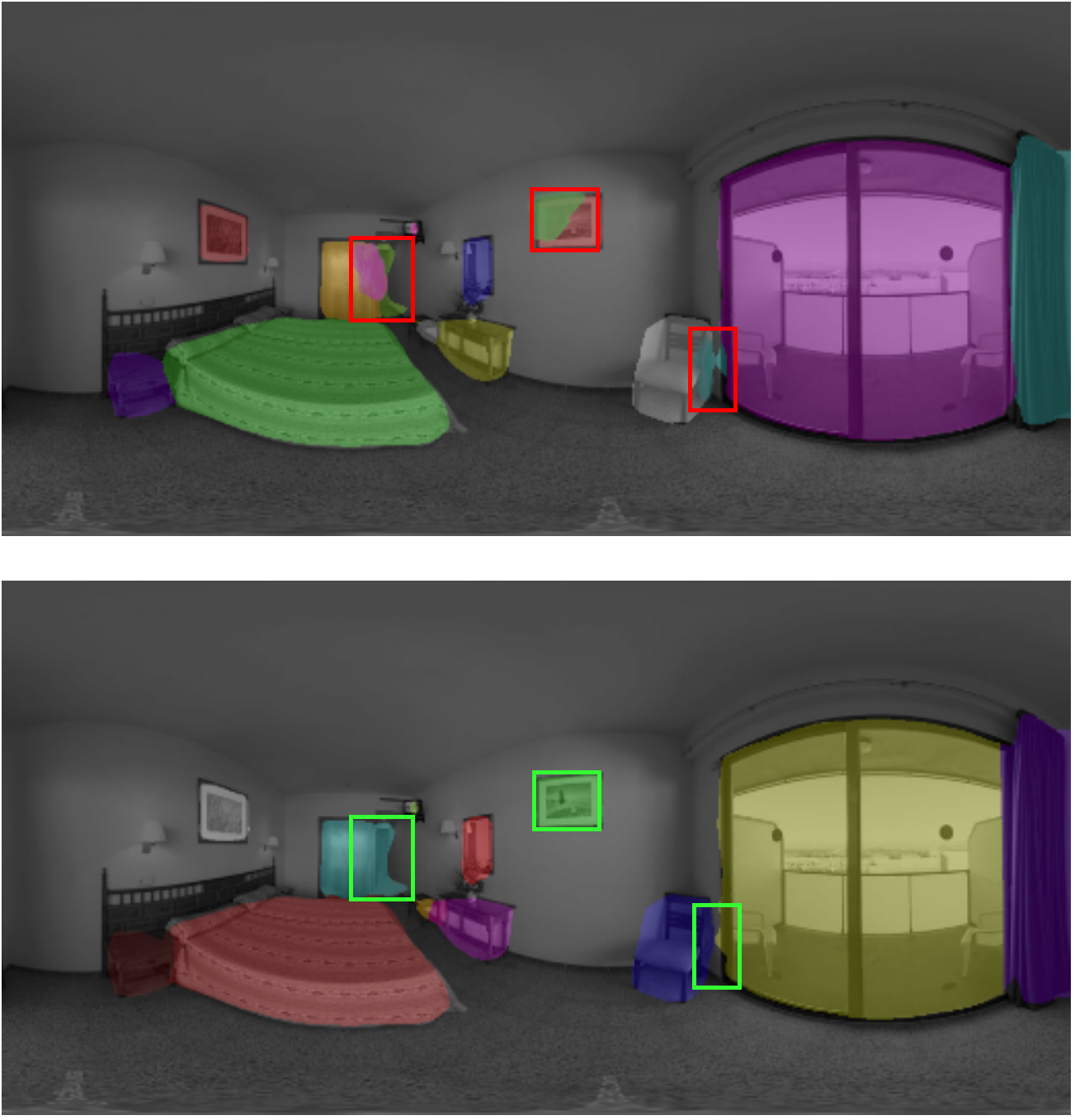}
   \end{minipage}\hfill
   \caption{\textbf{Instance segmentation post-processing} results. Top is initial semantic segmentation (output of CNN) and bottom is result of post-processing. Notice that apart from correctly differentiate among instances (highlighted in blue) it improves original segmentation (highlighted in red and green for failed and improved segmentation respectively).}
    \label{fig:instance}
\end{figure}

In this experiment we compare the semantic segmentation output of the network with the result of applying our instance segmentation method to create improved semantic segmentation maps. Although this is not the objective of the post-processing (the goal is to differentiate among different instances), it is evaluated to prove the influence that it can have in segmentation performance. Our intuition was that the instance segmentation method would imply an improvement to the initial segmentation maps because it gives higher confidence to detections, whose performance is clearly higher than segmentation's one in our model. Results, shown in Table~\ref{table:instance}, are very similar and lead us to conclude that the post-processing does not prove to be influential in this way. However, qualitative results support our intuitive idea by showing some clearly improving cases that are remarked in Figure~\ref{fig:instance}.

\begin{figure}
    \hspace{-0.5cm}
    \centering
    \includegraphics[scale=0.6]{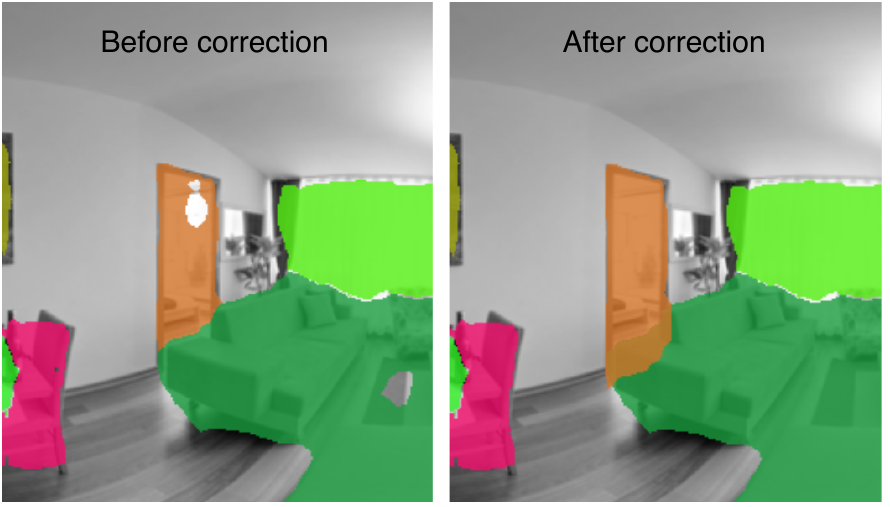}
    \caption{After combining the room layout with our segmentation masks, the model experiences a clear improvement as a whole. However, here we want to show a \textbf{failure case} where, when assuming that doors must reach the floor, we may have overlapping with other occluding objects in the image, damaging segmentation results but improving the door 3D localization.}
    \label{fig:roomMasks}
\end{figure}
%\vspace{-3mm}

%---------------- Tablas antes para que salgan bien colocadas

\begin{table*}[t]
\begin{center}
\small
\vspace{-1em}
\label{tab:vgg_all}
%\sffamily
\resizebox{\linewidth}{!}{
\begin{tabular}{x{95}|x{20} |yyyyyyyyyyyyyc}
  \ct{model} & mAP & \ct{bed} & \ct{painting} & \ct{table} & \ct{mirror} & \ct{window} & \ct{curtain} & \ct{chair} & \ct{light} & \ct{sofa} & \ct{door} & \ct{cabinet} & \ct{bedside} & \ct{tv} & \ct{shelf}  \\
  \hline\hline
  $*$ DPM & 29.4 & 35.2 & 56.0 & 21.6 & 19.2 & 21.8 & 29.5 & 26.0 & --- & 22.2 & 31.9 & --- & --- & 31.0 & ---  \\
   $*$ Deng \textit{et al.} & 68.7 & 76.3 & 68.0 & 73.6 & 58.7 & 62.6 & 69.5 & 68.0 & --- & 72.5 & 67.3 & --- & --- & 70.0 & ---  \\
    \hline
  \textit{PanoBlitzNet} StdConvs & 76.8 & 94.9 & \textbf{85.0} & \textbf{83.3} & 71.9 & \textbf{72.2} & 72.2 & 71.9 & 35.0 & \textbf{89.3} & \textbf{75.5} & \textbf{57.9} & 87.9 & 91.1 & 30.5  \\
  \textit{PanoBlitzNet} EquiConvs & \textbf{77.8} & \textbf{95.3} & 83.9 & 82.1 & \textbf{76.2} & 70.9 & \textbf{75.9} & \textbf{80.9} & \textbf{41.0} & 85.4 & 72.5 & 55.6 & \textbf{91.4} & \textbf{93.3} & \textbf{40.2}
\end{tabular}
}
\caption{\textbf{Object detection} results on SUN360 test set with our method \textit{Panoramic BlitzNet} using different convolutions (Standard vs. Equirectangular). $*$ Results trained and evaluated on a combination of datasets (including SUN360) by Deng \textit{et al.} \cite{Deng}}
\label{table:state_art_comparison_det}
\end{center}
\end{table*}

\begin{table*}[t]
\begin{center}
\small
\vspace{-1em}
\label{tab:vgg_all}
%\sffamily
\resizebox{\linewidth}{!}{
\begin{tabular}{x{95}|x{20} |yyyyyyyyyyyyyyc}
  \ct{model} & mIoU & \ct{backgrd.} & \ct{bed} & \ct{painting} & \ct{table} & \ct{mirror} & \ct{window} & \ct{curtain} & \ct{chair} & \ct{light} & \ct{sofa} & \ct{door} & \ct{cabinet} & \ct{bedside} & \ct{tv} & \ct{shelf}  \\
  \hline\hline
  PanoContext & 37.5 & 86.9  & \textbf{78.6} & 38.7 & 29.6 & 38.2 & 35.6 & --- & 09.6 & --- & 11.1 & 19.4 & 27.4 & 39.7 & 34.8 & ---  \\
    \hline
  \textit{PanoBlitzNet} StdConvs & 53.0 & 90.7 & 61.7 & 32.1 & \textbf{75.2} & \textbf{42.3} & \textbf{55.8} & \textbf{54.0} & 55.1 & \textbf{31.4} & \textbf{34.6} & 63.6 & 48.5 & \textbf{40.7} & 52.2 & 57.4  \\
  
  \textit{PanoBlitzNet} EquiConvs & \textbf{54.4} & \textbf{91.3} & 62.1 & \textbf{61.2} & 72.3 & 41.1 & 53.4 & 53.7 & \textbf{55.2} & 26.5 & 32.9 & \textbf{63.8} & \textbf{51.1} & 36.6 & \textbf{52.3} & \textbf{61.9} \\
  
%  \textit{PanoBlitzNet} EquiConvs$^{\dagger}$ & 54.5 & 91.3 & 63.7 & 59.2 & 72.1 & 40.3 & 54.3 & 55.0 & 55.6 & 29.1 & 30.1 & 63.1 & 51.4 & 36.1 & 53.6 & 62\\
  
%  \textit{PanoBlitzNet} EquiConvs$^{\ddagger}$ & \textbf{61.5} & 89.4 & \textbf{84.1} & \textbf{72.4} & \textbf{75.9} & \textbf{69.0} & \textbf{64.0} & \textbf{60.2} & \textbf{60.5} & \textbf{49.6} & \textbf{39.1} & 58.5 & \textbf{54.3} & \textbf{40.7} & 45.6 & 59.4
\end{tabular}
}
\caption{\textbf{Semantic segmentation} results on SUN360 extended test set with our proposed model \textit{Panoramic BlitzNet} using different convolutions (Standard vs. Equirectangular) and comparison with PanoContext~\cite{PanoContext}.}
%Here we compare PanoContext \cite{PanoContext} with the different steps of our method, thus being also considered as an ablation study of the proposed model. $^{\dagger}$ stands for our results after doing the instance segmentation post-processing and $^{\ddagger}$ stands for our final results after refining the masks with the room layout.}
\label{table:state_art_comparison_seg}
\end{center}
\end{table*}
\vspace{-1mm}
% ---------------------- FIN Tablas del estado del arte
Our approach proves to work well on several different scenes by correctly separating same category objects, that initially overlapped in semantic maps, into different instances. Limitations of the method can be seen when the network fails detecting an object, which is therefore not differentiated as an instance on the final map and when managing objects with complex shapes that can not be modelled with a gaussian distribution.

Here, we finally analyze the improvement over our segmentation masks by leveraging the contextual information of the room layout.
In our experiment, logical assumptions used for this refinement entail a significant improvement of up to $7.2\%$ mIoU with respect to the segmentation output of \textit{Panoramic BlitzNet} with StdConvs, achieving a final \textbf{mIoU = }$\textbf{60.3\%}$.
%Logical assumptions used for this refinement allow us to achieve a final $\textbf{mIoU = 61.5\%}$, which represents an improvement percentage of up to $7.0\%$. 
It should be noted that the classes that contribute most to this improvement are mirror, window and painting. However, while one would also expect a clear improvement in the door category, we have seen a drop in performance in some cases such as the one shown in Figure \ref{fig:roomMasks}, although it definitely has a positive effect on its location in the 3D room space. As already supported by this preliminary experiment, we propose a promising method to noticeably benefit 2D and 3D object recognition tasks from room layout knowledge, and encourage the idea that it is worth continuing to work in this direction.%As already supported by this preliminary experiment, we strongly believe on the use of these and further-studied assumptions from layout information as a promising method to noticeably benefit 2D and 3D object recognition tasks.

% ----- FIGURA para que salga bien
\vspace{-1mm}
% .--------------- FIN FIGURA 

% ----------------------------------------------------------
\subsection{Comparison with the State of the Art}
\noindent \textbf{Detection:} In Table~\ref{table:state_art_comparison_det} we show our detection results on the SUN360 extended dataset. Our Panoramic BliztNet with EquiConvs achieves very satisfactory results, with a global $\textbf{mAP = 77.8\%}$. For completeness, we include here the results of~\cite{Deng}, recent work on indoor panoramic object recognition with deep learning, together with their evaluation of DPM on panoramas. Our method achieves the best results in detection for all 10 common classes compared to them. It is worth noting that our approach achieves these results just training with $\sim400$ panoramas from the SUN360 dataset while they use additional panoramas to train their model. %confirm
Since their dataset is not public and no code is available, we report directly the results collected in \cite{Deng}.

\noindent \textbf{Segmentation:} Table~\ref{table:state_art_comparison_seg} summarizes the semantic segmentation results on the SUN360 extended dataset. A direct comparison is possible with the work of PanoContext \cite{PanoContext}. The results clearly show that our method significantly improves over the state of the art. In particular, we add three new object classes and boost $\textbf{mIoU = 54.4\%}$, which represents an improvement of $16.9\%$ over PanoContext's method.

\section{Conclusion}
%The proposed method provides a complete understanding of the main objects in an indoor scene from a single panoramic image. We perform object detection and semantic segmentation tasks adapting  BlitzNet architecture \cite{BlitzNet} to match the nature of the equirectangular image input and combine them to obtain instance segmentation masks. Finally, we post-process instance segmentation results to be placed, according to a series of geometrical and physical reasoning, inside the 3D model of the room. Quantitative analyses support that our method outperforms the state of the art by a large margin. Additionally, qualitative results show several 3D room models obtained with much less post-processing than the state of the art.
From a single panoramic image, we propose a method that provides a complete understanding of the main objects in an indoor scene. By managing the inherent characteristics and challenges that equirectangular panoramas involve, we outperform state of the art in addition to creating a more complete system, which not only obtains 2D detection and pixel-wise segmentation of objects but also places them into a 3D reconstruction of the room. Exploiting the advantages of having a wider field of view in indoor environments, this visual system becomes a promising key element for future autonomous mobile robots.
Future work includes the inclusion of instance segmentation predictions into the deep learning pipeline and a further study of the potential in combining layout recovery and object recognition tasks.

\addtolength{\textheight}{-8cm}   % This command serves to balance the column lengths

%%%%%%%%%%%%%%%%%%%%%%%%%%%%%%%%%%%%%%%%%%%%%%%%%%%%%%%%%%%%%%%%%%%%%%%%%%%%%%%%

%%%%%%%%%%%%%%%%%%%%%%%%%%%%%%%%%%%%%%%%%%%%%%%%%%%%%%%%%%%%%%%%%%%%%%%%%%%%%%%%

%%%%%%%%%%%%%%%%%%%%%%%%%%%%%%%%%%%%%%%%%%%%%%%%%%%%%%%%%%%%%%%%%%%%%%%%%%%%%%%%

\section*{ ACKNOWLEDGMENT}
{This work was supported by Projects RTI2018-096903-B-I00 (AEI/FEDER, UE), the Regional Council of Bourgogne Franche-
Comte (2017-9201AAO048S01342) and Mobility City.}

\clearpage
%%%%%%%%%%%%%%%%%%%%%%%%%%%%%%%%%%%%%%%%%%%%%%%%%%%%%%%%%%%%%%%%%%%%%%%%%%%%%%%%

{%\small
\bibliographystyle{IEEEtran}
\bibliography{egbib}
}

% \begin{thebibliography}{99}

% \bibitem{c1} C. J. Kaufman, Rocky Mountain Research Lab., Boulder, CO, private communication, May 1995.

% \end{thebibliography}

\end{document}